\pdfoutput=1

\documentclass[11pt]{article}

\usepackage{acl}

\usepackage{times}
\usepackage{latexsym}
\usepackage{graphicx} 
\usepackage{booktabs}
\usepackage{hyperref}
\usepackage{makecell} 

\usepackage[T1]{fontenc}

\usepackage[utf8]{inputenc}

\usepackage{microtype}

\title{\textsc{HybriDialogue}: An Information-Seeking Dialogue Dataset\\ Grounded on Tabular and Textual Data}

\author{Kai Nakamura\textsuperscript{1}, Sharon Levy\textsuperscript{2}, Yi-Lin Tuan\textsuperscript{2}, Wenhu Chen\textsuperscript{3}, William Yang Wang\textsuperscript{2} \\
    \textsuperscript{1} California Institute of Technology \\
    \textsuperscript{2} University of California, Santa Barbara \\
 \textsuperscript{3} University of Waterloo, Vector Institute \\
  \texttt{kai.nakamura42@gmail.com}\\ 
  \texttt{\{sharonlevy,ytuan,william\}@cs.ucsb.edu} \\
\texttt{wenhuchen@uwaterloo.ca}}

\begin{document}
\maketitle
\begin{abstract}
A pressing challenge in current dialogue systems is to successfully converse with users on topics with information distributed across different modalities. Previous work in multiturn dialogue systems has primarily focused on either text or table information. In more realistic scenarios, having a joint understanding of both is critical as knowledge is typically distributed over both unstructured and structured forms. We present a new dialogue dataset, \textsc{HybriDialogue}, which consists of crowdsourced natural conversations grounded on both Wikipedia text and tables. The conversations are created through the decomposition of complex multihop questions into simple, realistic multiturn dialogue interactions. We propose retrieval, system state tracking, and dialogue response generation tasks for our dataset and conduct baseline experiments for each. Our results show that there is still ample opportunity for improvement, demonstrating the importance of building stronger dialogue systems that can reason over the complex setting of information-seeking dialogue grounded on tables and text. 


\end{abstract}

\section{Introduction}

When creating dialogue systems, researchers strive to enable fluent free-text interactions with users on a number of topics. These systems can be utilized to navigate users over the vast amount of online content to answer the user's question. Current systems may search for information within text passages. However, knowledge comes in many forms other than text. The ability to understand multiple knowledge forms is critical in developing more general-purpose and realistic conversational models. Tables often convey information that cannot be efficiently captured via text, such as structured relational representations between multiple entities across different categories ~\cite{chen2019tabfact,chen-etal-2020-hybridqa,herzig-etal-2020-tapas}. On the other hand, text may contain more detailed information regarding a specific entity. Thus, dialogue systems must be able to effectively incorporate and reason across both modalities to yield the best performance in the real world. 

While there are several existing datasets targeted at dialogue systems~\cite{dinan2018wizard,budzianowski-etal-2018-multiwoz,eric-etal-2017-key,zhou-etal-2018-dataset}, these are limited to either table-only or text-only information sources. As a result, current dialogue systems may fail to respond correctly in situations that require combined tabular and textual knowledge.


\begin{figure*}[t]
  \centering
  \includegraphics[width=\linewidth]{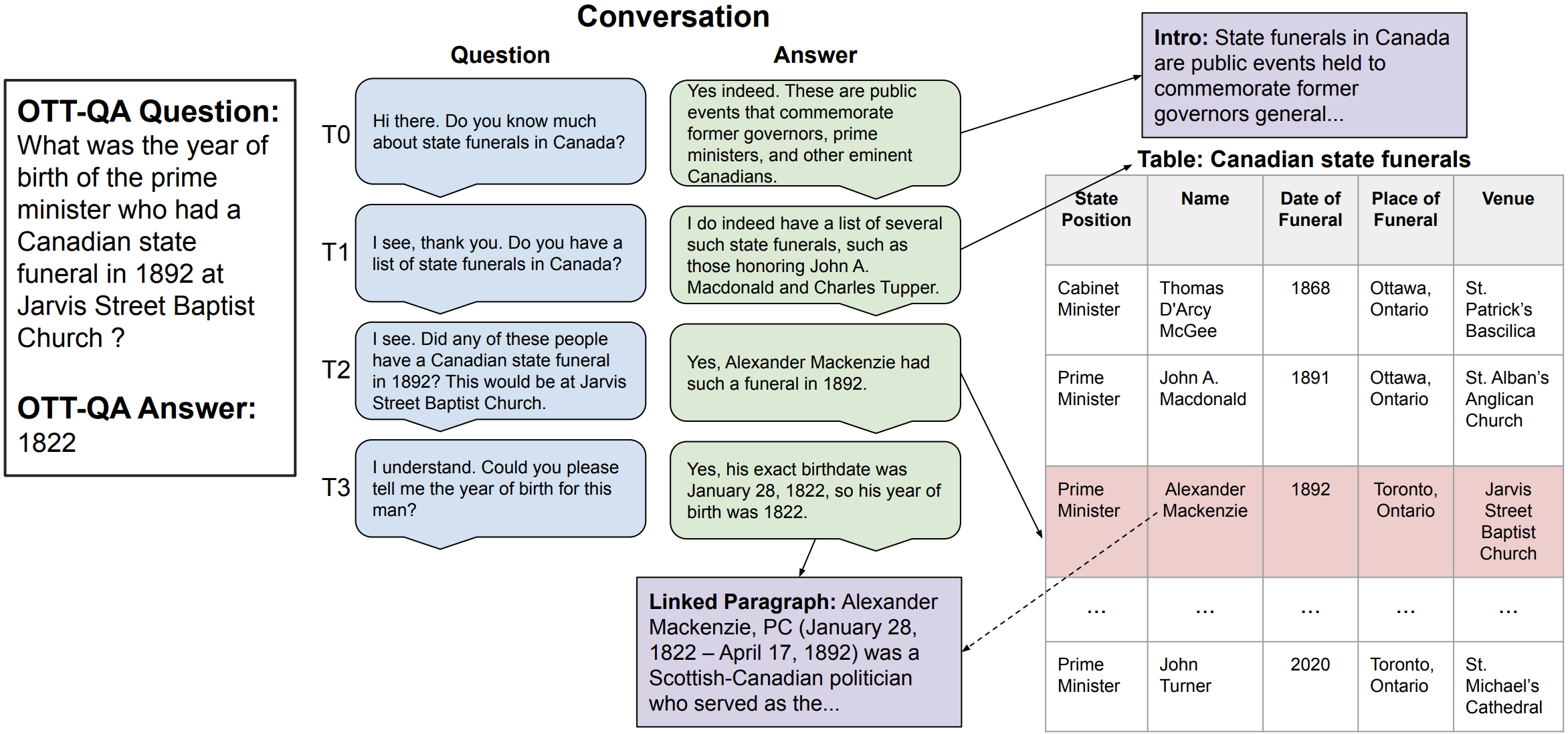}
  \caption{Overview of a sample from \textsc{HybriDialogue}, where each conversation is created from a decomposed multihop question-answer pair. T0,...,T3 represent turns in the dialogue and consist of a single question and answer pair. The solid arrows represent the reference (e.g., row or intro paragraph) utilized to retrieve the correct answer in each turn. The dashed arrow represents a paragraph linked from a table cell.}
\end{figure*}\label{fig:overview}

To advance the current state of dialogue systems, we create \textsc{HybriDialogue} \footnote{\url{https://github.com/entitize/HybridDialogue}}. Our dataset is an information-seeking dialogue dataset grounded on structured and unstructured knowledge from tables and text. \textsc{HybriDialogue}, or \textsc{HyDi}, is constructed by decomposing the complex and artificial multihop questions in OTT-QA~\cite{chen2020open} which may not reflect real-life queries. We transform these into a series of simple and more realistic intermediate questions regarding tables and text that lead to and eventually answer the multihop question. \textsc{HybriDialogue} contains conversations written by crowdsourced workers in a free-flowing and natural dialogue structure that answer these simpler questions and the complex question as well.   We provide an example dialogue from our dataset in Figure \ref{fig:overview}. We also propose several tasks for \textsc{HybriDialogue} that illustrate the usage of an information-seeking dialogue system trained on the dataset. These tasks include retrieval, system state tracking, and dialogue generation. Together, they demonstrate the challenges with respect to dialogue systems and the necessity for a dataset such as \textsc{HybriDialogue} to further research in this space.


Our contributions are as follows:
\begin{itemize}
  \item We create a novel dialogue dataset consisting of 4800+ samples of conversations that require reasoning over both tables and text. 
  \item We decompose the overly-complex multihop questions from an existing dataset into more realistic intermediate question-answer pairs and formulate these in the dialogue setting.
  
  
  \item We propose system state tracking, dialogue generation, and retrieval tasks for our dataset. Our baseline experiments demonstrate opportunities to improve current state-of-the-art models in these various tasks and the overall information-seeking dialogue setting.
\end{itemize}

\section{Related Work}
Related work in the space of dialogue-based question-answering can be split into two areas: question-answering systems and information-grounded dialogue. We provide a comparison of the related datasets in Table \ref{tab:my-table} and analyze these datasets below.


\begin{table}[t!]
\begin{minipage}{0.5\textwidth}
\small
\centering
\setlength{\tabcolsep}{5.5pt}
\begin{tabular}{l|l|l|l}
\toprule
\textbf{Dataset} & \textbf{Dialogue} & \textbf{Turns} & \textbf{Modality}  \\
\midrule[0.5pt]
CoQA          & 8K                 & 127K        & Text                                  \\
Natural Questions          & 0                 & 323K        & Text                                  \\
Hybrid-QA      & 0                  & 7k           & Table/Text                       \\
OTT-QA         & 0                  & 45K          & Table/Text                        \\
SQA            & 6.6K               & 17.5K        & Table                      \\
ShARC          & 948                & 32K            & Text                                            \\
DoQA           & 2.4K               & 10.9K        & Text                                                     \\
RecipeQA       & 0                  & 36K          & Image/Text                             \\\midrule[0.5pt]
KVRET & 3K  & 12.7K & Table\\
MultiWOZ & 10.4K & 113.6K & Table\\
WoW & 22.3K  & 101K & Text\\
Topical-Chat & 10.8K & 235.4K & Text\\
CMU\_DoG & 4.2K & 130K & Text\\\midrule[0.5pt]
\textbf{\textsc{HybriDialogue}} & 4.8K               & 22.5K        & Table/Text                 \\
\bottomrule
    \end{tabular}
        \caption{Comparison of \textsc{HybriDialogue} and other\\ dialogue and question-answering datasets. For question\\-answering datasets, turns refers to question-answer 
        \\ pairs. For ShARC, dialogues refers to dialogue trees.}
    \label{tab:my-table}
\end{minipage}    
\end{table}

\paragraph{Question-Answering}
As question-answering (QA) is one of the long-established NLP tasks, there are numerous existing datasets related to this task. 
Recently, QA datasets have been incorporating new modalities. The Recipe-QA~\cite{yagcioglu-etal-2018-recipeqa} dataset is comprised of question-answer pairs targeted at both image and text. OTT-QA~\cite{chen2020open} and Hybrid-QA~\cite{chen-etal-2020-hybridqa} both contain complex multihop questions with answers appearing in both text and tabular formats. 
Several datasets are also targeted at the open-domain question-answering task such as TriviaQA, HotPotQA, and Natural Questions ~\cite{joshi-etal-2017-triviaqa,yang-etal-2018-hotpotqa,kwiatkowski-etal-2019-natural}.
While single-turn question-answering is valuable, the dialogue setting is more interesting as it proposes many new challenges, such as requiring conversational context, reasoning, and naturalness. 


\paragraph{Conversational Question-Answering}
Several question-answering datasets contain question and answer pairs within a conversational structure. CoQA~\cite{reddy-etal-2019-coqa} and DoQA~\cite{campos-etal-2020-doqa} both contain dialogues grounded with knowledge from Wikipedia pages, FAQ pairs, and other domains. ShARC~\cite{saeidi-etal-2018-interpretation} employs a decomposition strategy where the task is to ask follow-up questions to understand the user's background when answering the original question. However, ShARC is limited to rule-based reasoning and `yes' or `no' answer types. SQA~\cite{iyyer-etal-2017-search} provides a tabular-type dataset, consisting of the decomposition of WikiTable questions. Each decomposed answer is related to a cell or column of cells in a particular table. 
In these datasets, knowledge is limited to a single modality.

In comparison, our dataset poses a more challenging yet realistic setting, where knowledge over structured tables and unstructured text is required to provide reasonable answers to the conversational questions. 
While the previous datasets contain samples written in a conversational structure, the answers are not necessarily presented in this way; they will instead formulate simple and short answers that do not emulate a human dialogue. 
 Our dataset, therefore, extends conversational question-answering and falls into the dialogue space.  \textsc{HybriDialogue} contains natural dialogues with strongly related question-answer pair interactions whose answers are longer than the exact answer string. This models real-world occurrences in which a person wants to ask follow-up questions after their initial question has been answered.

\paragraph{Dialogue Generation}
Among the dialogue datasets that leverage structured (tables and knowledge graphs) knowledge, some~\cite{ghazvininejad2018knowledge,zhou2018commonsense} use conversational data from Twitter or Reddit and contain dialogues relying on external knowledge graphs such as Freebase~\cite{bollacker2008freebase} or ConceptNet~\cite{speer2017conceptnet}.
On the other hand, OpenDialKG~\cite{moon2019opendialkg}, DuConv~\cite{wu2019proactive}, DyKGChat~\cite{tuan2019dykgchat}, and KdConv~\cite{zhou2020kdconv} collect conversations that are explicitly related to the paired external knowledge graphs.
Other related work revolves around task-oriented dialogues that are grounded on tables. For example, KVRET~\cite{eric-etal-2017-key} and MultiWOZ~\cite{budzianowski-etal-2018-multiwoz,ramadan2018large,eric2019multiwoz,zang2020multiwoz} provide tables that require an assistant to interact with users and complete a task.

Dialogue datasets that are grounded on unstructured knowledge include CMU\_DoG~\cite{zhou-etal-2018-dataset}, which is composed of conversations regarding popular movies and their corresponding simplified Wikipedia articles.
On the other hand, Wizard-of-Wikipedia (WoW)~\cite{dinan2018wizard} and Topical-Chat~\cite{Gopalakrishnan2019} simulate the human-human conversations through Wizard-Apprentice, in which the apprentice tries to learn information from the wizard.
Our proposed task shares a similar idea with Wizard-of-Wikipedia and Topical-Chat in terms of asymmetric information among participants. However, we focus more on information-seeking dialogues grounded on both structured and unstructured knowledge, which provides abundant and heterogeneous information, and requires joint reasoning capabilities using both modalities.


\begin{figure}[t]
  \centering
  \includegraphics[width=\linewidth]{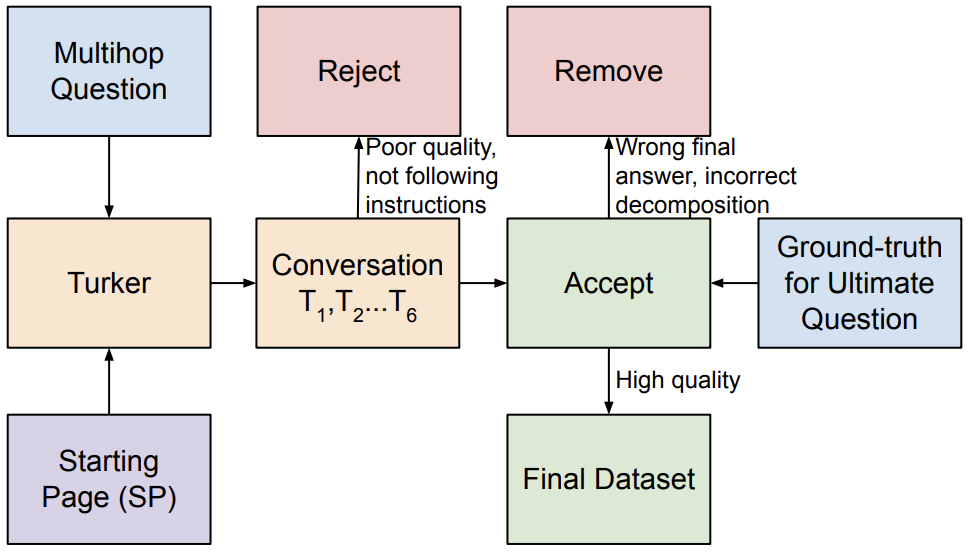}
  \caption{Overview of the dataset collection process, including the validation steps.}\label{fig:collection}
\end{figure}

\section{Dataset Creation}\label{sec:creation}


\subsection{Crowdsourcing Instructions}
Given a multihop question from OTT-QA, crowdsourced workers (Turkers) from Amazon Mechanical Turk~\cite{mechanicalturk} were asked to decompose it into a series of simpler intermediate questions and answers to formulate a simulated conversation in English. 
\footnote{\url{https://confident-jennings-6a2f67.netlify.app/plaid_interfaces/examples/1a_example_1.html}}
As opposed to datasets such as Wizard of Wikipedia \cite{dinan2018wizard} that are more open-ended, our annotators have a specific goal in mind: to answer an original complex question. By utilizing a single annotator to represent both sides, we keep the flow of the dialogue consistent and natural as it converges to the final answer. The usage of two annotators for our specific task comes with the risk of having one user diverge and reduce the chance of reaching the correct final answer.

\begin{figure}[t]
  \centering
  \includegraphics[width=0.6\linewidth]{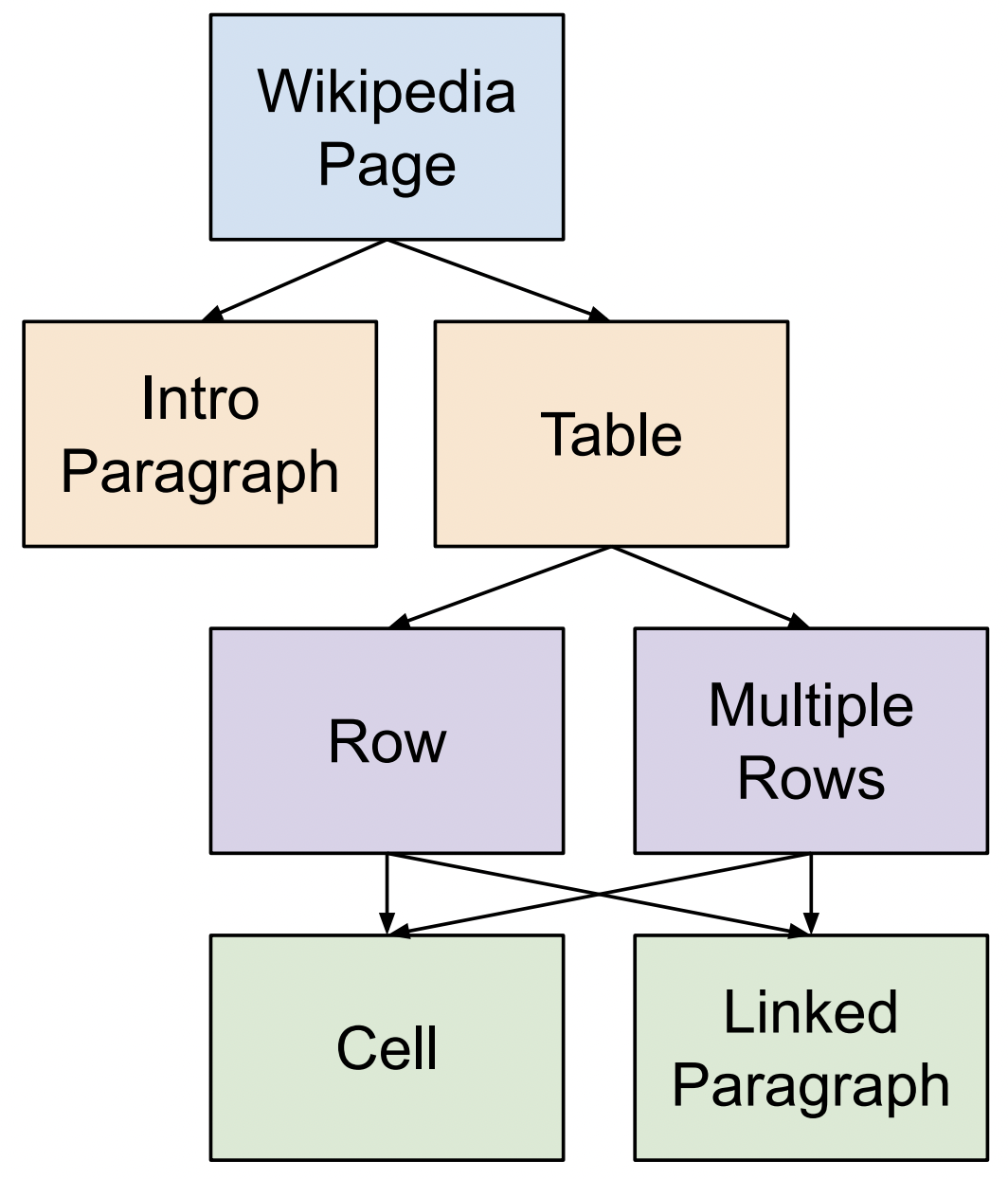}
  \caption{Overview of the reference pool graph, indicating which reference candidates are added to the pool given the current available references.}\label{fig:state_transition}
\end{figure}

We refer to the multihop question from OTT-QA as the ``ultimate question''. Turkers are instructed as follows: ``In this task, you will engage in a dialogue with yourself. You will act as two characters: the seeker and the expert. At the top of the page, you are given the Ultimate Question. The seeker wants to know the answer to the ultimate question. However, directly asking this ultimate question is too complex. Thus, the seeker needs to decompose (break down) this complex question into a sequence of simple questions, which the expert will answer using a database.''
To further emphasize the naturalness of the dataset, Turkers were encouraged to ask questions that required understanding the conversation history context, such as through co-referencing. For example, Turkers used proper nouns with pronouns and indirect references such that they logically refer to their antecedents. 
An example conversation is demonstrated in Figure \ref{fig:overview} and an overview of the dataset collection process is shown in Figure \ref{fig:collection}. 

\subsection{Task Definitions}
A conversation is composed of a sequence of turns. Each conversation consists of a minimum of 4 turns and a maximum of 6 turns. This limitation is specified to ensure that Turkers are thoroughly decomposing each complex question and the conversations do not go off on tangents. Each turn $T$ acts as a piece of the decomposition of the ultimate question. The i-th turn $T_i$ consists of a natural language question $Q_i$, a natural language answer $A_i$, a reference $R_i$ from an English Wikipedia page, and an available reference pool set ${RP}_i$. The Turker provides $Q_i$, $A_i$, and selects a particular $R_i$ from the set $RP_i$. $R_i$ can be considered the evidence required to generate $A_i$ given the question $Q_i$. The reference pool $RP_i$ contains different types of references including the (linked) paragraph, a (whole) table, a single inner table row, multiple inner table rows, or a single cell. 

We differentiate between multiple rows and the whole table in order to obtain a more specific source for the information. 
For example, the question "Do you have a list of Steve's accomplishments?" requires a Table response as the answer contains a summary of the table.
On the other hand, the question "Did he ever compete in the Grand Prix event type?" requires a selection of specific
rows of some table.
In order to enforce the naturalness and moderate the difficulty of questions, we restricted ${RP}_i$ based on $RP_{i - 1}$ and $R_{i - 1}$. In other words, the type of questions that the Turker could ask were restricted to the references enabled from previous selections. In the Turker interface, $RP_0$ is restricted to the intro paragraph and any whole table references in a provided starting page. We illustrate how reference candidates are added to the reference pool in Figure \ref{fig:state_transition}.

\begin{table}[t]
\centering
\makebox[\linewidth]{
\begin{tabular}{{l}|{l}}
\toprule
Dataset Statistics \\
\hline
\# Train Dialogues          & 4359  \\
\# Development Dialogues          & 242  \\ 
\# Test Dialogues          & 243  \\ 
\# Turns (QA pairs)       & 21070 \\
Avg Turns per Dialogue  & 4.34 \\
\# Wikipedia Pages    & 2919 \\
Avg \# words per question & 10    \\
Avg \# words per answer   & 12.9  \\ 
\# Table selections              & 4975  \\ 
\# Row selections                & 6769  \\
\# Cell selections               & 1830  \\ 
\# (Linked) paragraph selections  & 3337  \\
\# Intro selections & 7131  \\
\# Unique decompositions   & 267   \\

\bottomrule
 \end{tabular}
 }
\caption{\textsc{HybriDialogue} dataset statistics. }\label{tab:stats}
\end{table}

\begin{figure*}[t]
  \centering
  \includegraphics[width=\linewidth]{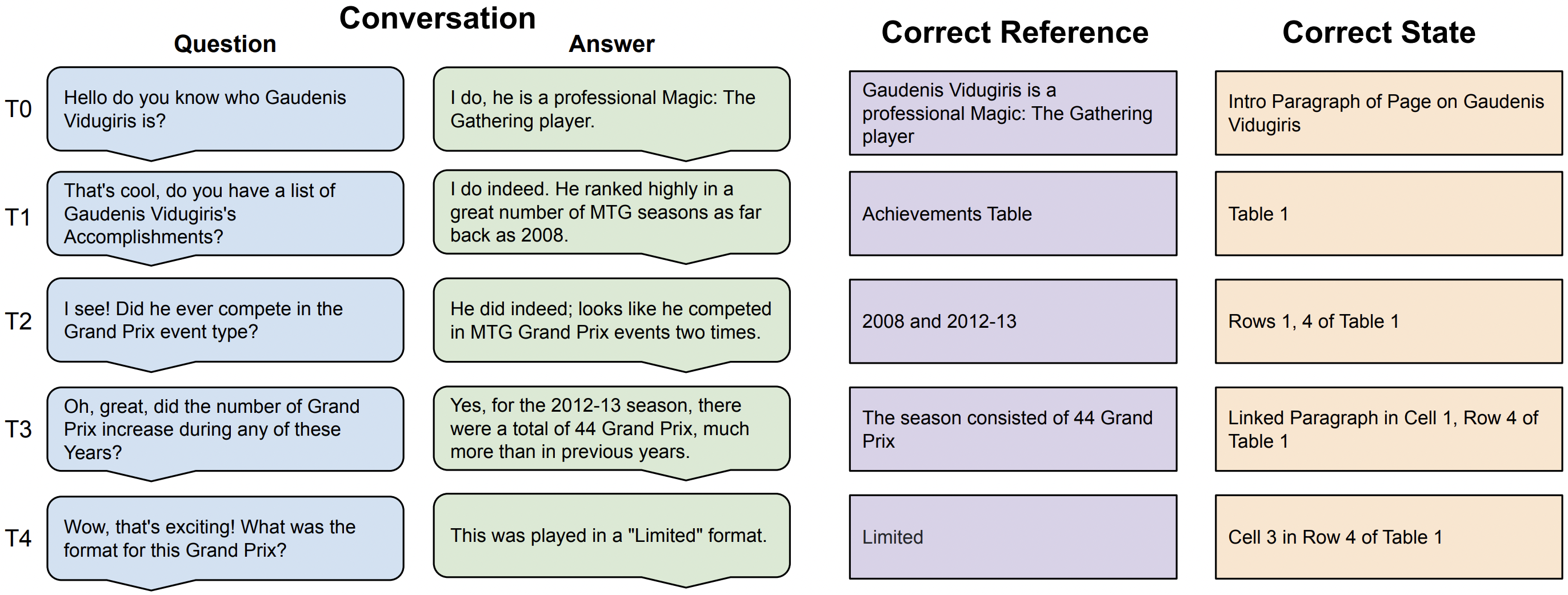}
  \caption{Overview of the state tracking experiment. For each question in a conversation turn, there is a correct reference and corresponding state (e.g., row, linked paragraph) to select when answering the question.}\label{fig:statetracking}
\end{figure*}


\subsection{Validation} 

To ensure high-quality samples, we conducted various filtering steps. Rejections were made due to the Turker not following the instructions at all or having poor-quality conversations. For example, if the Turker purposefully copy and pasted unrelated paragraphs of texts, repeated the same questions multiple times, used unrelated references, or utilized a single reference throughout the entire conversation, we automatically rejected it. Turkers were paid an average of \$1.1 per conversation. Completing a conversation took the worker an average of 5 minutes, which translates to an average of \$13.2 per hour. 
In some cases, we gave bonuses to Turkers who consistently submitted high-quality results. After final verification of the accepted HITs, we obtained a final dataset consisting of 4,844 conversations. The statistics of the dataset are shown in Table \ref{tab:stats}.

We conducted additional filtering to further enhance the dataset quality. Utilizing gold answers obtained from the source OTT-QA dataset, we checked if the final answer appeared as a substring in Turker’s conversation. If it did, we auto-approved the conversation. For the remaining questions, we manually reviewed them. We approved conversations that had the correct answer but in a different format (e.g., September 1, 2021, instead of 9/1/21). In some cases, Turkers provided their own decomposition or their own ultimate question and decomposition, so they did not obtain the final answer provided by OTT-QA. 
In these cases, if the conversation was both accurate and had good quality, we accepted it. 

\section{Tasks and Baseline Models}
We outline three different tasks in the following sections: retrieval, system state tracking, and dialogue generation. Together, these tasks formulate a pipeline dialogue system grounded on both structured and unstructured knowledge from tables and text. The first step of the system is to \textbf{retrieve} the correct Wikipedia reference given the first question in the dialogue. As the conversation continues, the system must be able to \textbf{track the state} of the conversation in order to obtain the correct information from the Wikipedia reference for the user. Finally, the system will need to \textbf{generate a natural conversational response} to communicate with the user at each turn. Thus, following each of these tasks in order simulates the pipeline system with our dataset. We describe each of these tasks and their respective models in detail below.

\begin{figure*}[t]
  \centering
  \includegraphics[width=\linewidth]{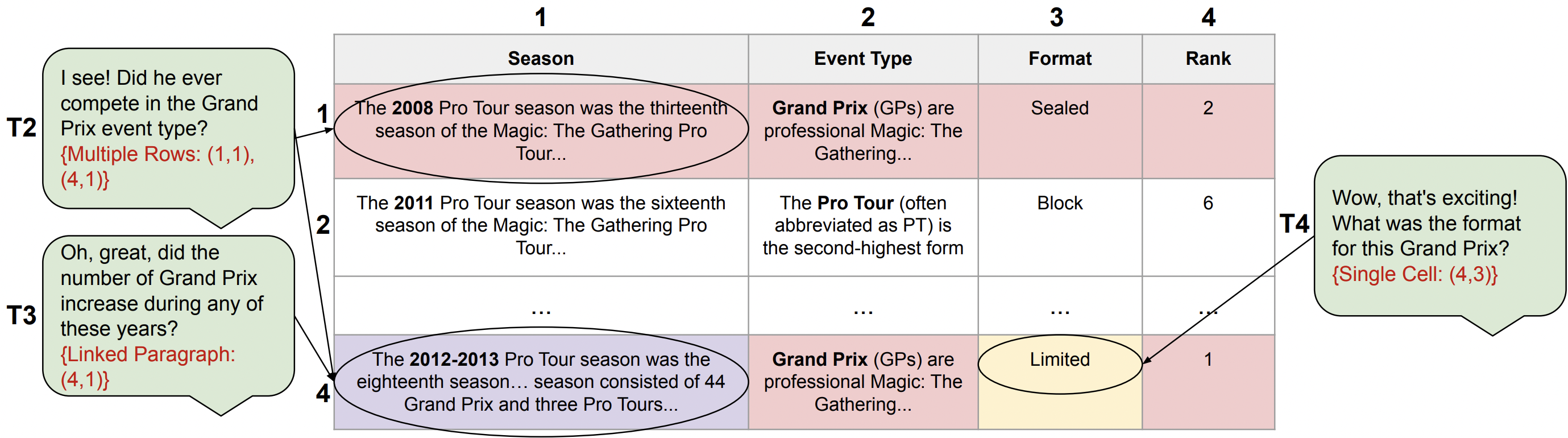}
  \caption{System state tracking with the TaPas model. Single rows and multiple rows are mapped to single cells and linked paragraphs are mapped to their respective cells in the original table in order to adapt to TaPas.}\label{fig:tapas}
\end{figure*}

\begin{figure}[t]
  \centering
  \includegraphics[width=\linewidth]{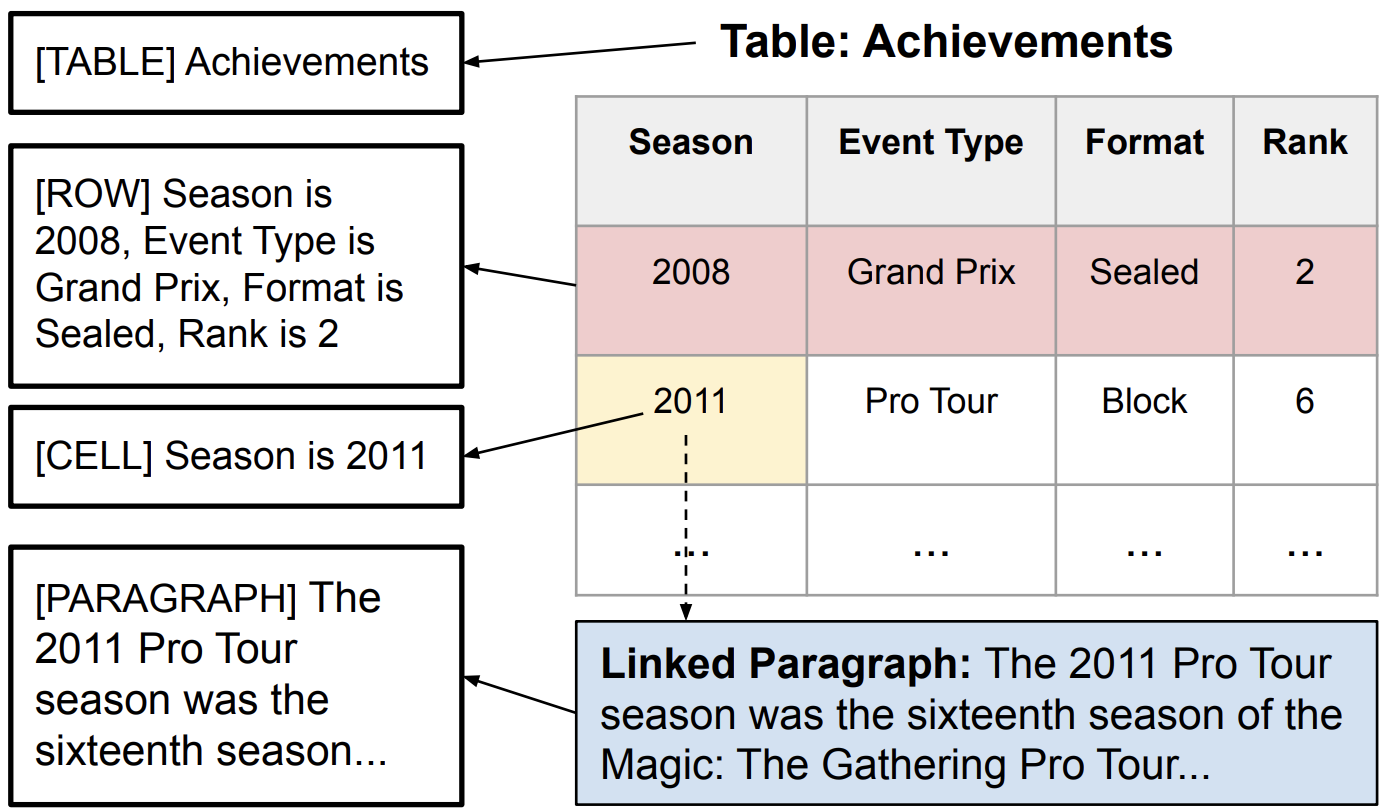}
  \caption{Table, row, cell, and paragraph flattening for input to the SentenceBERT and DialoGPT models.}\label{fig:sbert_dialogpt}
\end{figure}

\subsection{Retrieval}

The retrieval experiment is run for each $T_0$ of each conversation. Given the first question of the conversation $Q_0$, the model must predict the correct reference $R_0$. First questions discuss information that is either in a table or an intro paragraph; so the candidate space contains all intro paragraphs and tables in the dataset. 
The purpose of the retrieval experiment is to get a baseline of how well we are able to predict the table or page the subsequent conversation will be based upon, given the first query. The references that are utilized in the subsequent conversation are on the same page as the selected intro paragraph or table. 
For our baseline, we run the Okapi BM25 retriever \cite{rank_bm25} on the entire dataset over all candidates and first turn queries. BM25 is a standard document retrieval model that uses keyword-matching techniques to rank documents. 



\subsection{System State Tracking}

Previous work in dialogue systems focuses on the task of belief state tracking, which aims to determine the user's goal or the current state of the conversation at each turn in the dialogue ~\cite{mrksic-etal-2017-neural,ren-etal-2018-towards}. Inspired by work in belief state tracking, we propose the task of system state tracking in an information-seeking dialogue system. The task is framed similarly to belief state tracking, where a model attempts to classify the current state in the conversation at each turn. However, the ``state'' in our proposed task is modeled as a reference location from the current reference pool. As such, the task is formulated as using the information from the existing conversation and current question to determine the state of the conversation and choose which reference to utilize to create an answer. The reference types considered in this experiment are single cell, linked paragraph, inner table row, and multiple inner table rows. The implementation of system state tracking increases the interpretability and explainability of the system by determining the understanding of the user's question and discovering the point in the conversation in which the model is incorrectly interpreting the user's question. This, in turn, can help us understand the types of errors the model is prone to and allow us to work towards increasing the robustness of the model regarding these errors.

The system state tracking process is visualized in Figure \ref{fig:statetracking}.
We perform system state tracking for all turns in each dialogue except the first turn.
Given the history of the conversation $H_i$, we predict the correct reference $R_i$. $H_i$ consists of turns $T_1 ... T_{i-1}$, the current query $Q_i$, and the candidate references $RP_i$.
Thus, the goal is to determine the correct reference $R_i$ at the specific turn in the dialogue, given the dialogue history. 
We utilize SentenceBERT \cite{reimers-2019-sentence-bert} and TaPas~\cite{herzig-etal-2020-tapas} as baselines for the experiment. 

\begin{table*}[t]
\centering
\begin{tabular}{l|l|c|c|c}\toprule[1pt]
\bf Task & \bf Model & \textbf{\# Samples}  & \bf MRR@10 & \bf MAP   \\
\midrule[0.5pt]
Retrieval & BM25 & 4844 & 0.427 & 0.427  \\
\midrule[0.5pt]
System State Tracking & SentenceBERT & 636 & 0.603 & 0.600  \\
& TaPas  & 636 & \textbf{0.689} & \textbf{0.634}  \\
\bottomrule[1pt]
 \end{tabular}
\caption{The results of the retrieval and system state tracking experiments. }\label{tab:retrieval_statetracking}
\end{table*}

\begin{table}[t]
\centering
\begin{tabular}{l|c|c|c}\toprule[1pt]
\bf Reference  & \bf MRR@10 & \bf MAP & \bf Count \\
\midrule[0.5pt]
Cell  & 0.384 & 0.395 & 108  \\
Paragraph   & 0.599 & 0.606 & 124 \\
Row  & 0.782 & \textbf{0.786} & 338 \\
Multi-row  & \textbf{0.881} & 0.292 & 66  \\
\bottomrule[1pt]
 \end{tabular}
\caption{System state tracking results split by reference type for the TaPas model.} \label{tab:statetracking_analysis}
\end{table}

\paragraph{SentenceBERT}

 We utilize the sentence transformer and the triplet-loss configuration as described in equation \ref{eq:triplet}. We minimize the difference between the correct candidate $R_i$ and context $H_i$ while maximizing the difference between every incorrect candidate $W$ and $H_i$. We create samples for each $W \in {RP}_i$ where $W \neq R_i$. ($RP_i$ is the reference pool). $k$ is some fixed margin.
 
\begin{equation} \label{eq:triplet}
loss = max(||H_i - R_i|| - ||H_i - W|| + k, 0) 
\end{equation}

To allow SentenceBERT to process the data, we flatten the references and prepend a special token to provide information about the type of candidate it is. This process is visualized in Figure \ref{fig:sbert_dialogpt}. 

\paragraph{TaPas}
We additionally utilize the TaPas model for system state tracking. TaPas is a BERT-based question-answering model for tabular data. We use the TaPas model that has been fine-tuned on the SQA dataset, which enables sequential question-answering in a conversational nature. As the model performs only cell selection, we adapt TaPas towards this setting. We do not need to pre-process the data differently for cell selection as TaPas already performs the cell selection task. We place linked paragraphs in their respective cells within a table to accommodate cell selection in this setting. For row and multi-row selection, we pre-process the data by choosing one cell from the row as the correct answer. This is done by finding the cell with the highest text similarity to the ground truth answer at that turn. Therefore, each row will have a single cell associated with it during fine-tuning. We visualize the state tracking experiment with TaPas in Figure \ref{fig:tapas}. For our experiments, we fine-tuned the TaPas model with our pre-processed training set.

\subsection{Dialogue Generation}

We conduct experiments on dialogue response generation to look into the dataset's expressivity for real-world dialogue scenarios.
We fine-tuned a pre-trained DialoGPT model~\cite{zhang2020dialogpt} by minimizing the negative log-likelihood with two input settings.
$Q_i$, $A_i$, and $R_i$ are defined as the question, answer, and reference at the i-th turn, respectively.
First, we only take the dialogue history as the input without knowledge content and predict the following natural language response.
The format (DialoGPT-noR) is described as:
\begin{equation}
    \{Q_1,A_1,...,Q_i,A_i,Q_{i+1}\} \mapsto A_{i+1} 
\end{equation}
Second, we flatten the references and concatenate the dialogue history as the input and predict the following natural language response. The references are flattened in the process seen in Figure \ref{fig:sbert_dialogpt}.
The format (DialoGPT) is:
\begin{equation}
    \{R_1,Q_1,A_1,...,R_{i+1},Q_{i+1}\} \mapsto A_{i+1} 
\end{equation}
The two settings enable us to validate how much information the references provide for response construction.

\begin{table}[t]
    \centering
    \begin{tabular}{l|c|c}\toprule[1pt]
        \bf Method & \bf SacreBLEU & \bf BERTscore \\\midrule[0.5pt]
        DialoGPT-noR & 14.72 & 0.8875\\
        DialoGPT & 21.63 & 0.8901\\\bottomrule[1pt]
    \end{tabular}
    \caption{The results of dialogue generation experiments on \textsc{HybriDialogue} dataset.}
    \label{tab:dialog-gen}
\end{table}

\begin{table*}[t]\small
    \centering
    \begin{tabular}{l|p{7cm}|p{5cm}}\toprule[1pt]
        \bf Error Type & \bf Input & \bf Output \\\midrule[0.5pt]
         Incoherent & [TABLE] Best-selling physical singles -- 7–7.9 million copies $ $; [QUERY] Can you give me a list ... [PARAGRAPH] .... Known for her emotive mezzo-soprano voice, Morissette began her career in Canada \dots; [QUERY] What is the vocal range of this singer?  & DialoGPT: Alanis Nadine Morissette is a Canadian-American singer, songwriter, and actress.
\newline GT: Alanis Morissette has a vocal range of mezzo-soprano. \\\midrule[0.5pt] 
        Non-fluent & [ROW] Year is 1985 ; Song is La det swinge ; Artist is Bobbysocks ; Position is 1st ; Points is 123... [QUERY] Do you know what song they performed to win? & DialoGPT: Yes, they performed to win the song La det swinge.\newline GT: Yeah, they did La det swinge.\\\midrule[0.5pt]
        Unfaithful & [PARAGRAPH] Immigration to Spain...in 2005 alone, the immigrant population of Spain increased by 700,000 people.; [QUERY] when did the immigrant population of spain increase by 700,000 people? & DialoGPT: The immigrant population increased by 700,000 in 1998.\newline GT: the immigrant population increased by that number in 2005\\
        \bottomrule[1pt]
    \end{tabular}
    \caption{The error types observed in dialogue generation on  \textsc{HybriDialogue}. (GT: ground truth)}
    \label{tab:dialog-gen-errors}
\end{table*}

\section{Experiments}
\subsection{Retrieval}
As retrieval is the first step in the information-seeking dialogue pipeline, we need to ensure that information from the correct Wikipedia page is retrieved to determine whether the first question and any following questions will be answerable. We evaluate our retrieval model with MRR@10 (Mean Reciprocal Rank @10). Table \ref{tab:retrieval_statetracking} shows our results, where BM25 achieves an MRR@10 score of 0.427 for retrieving the correct candidate.


\subsection{System State Tracking}
\paragraph{Evaluation}
To evaluate the SentenceBERT and TaPas predictions, we calculate MRR@10 (Mean Reciprocal Rank @10) and MAP (Mean Average Precision). Each model produces scores for the candidate references for a question. These scores are sorted into a ranked list, and the correct references are identified in this list. We then calculate MRR and MAP values with respect to the ranking of the correct reference in the ranked list.


When evaluating the TaPas model, we consider the highest-ranking cell from the ground truth row correct during test time. This simulates a more realistic setting by allowing any cell within the row to be correct.

\paragraph{Results}

The results of our experiments with TaPas and SentenceBERT are shown in Table \ref{tab:retrieval_statetracking}. Our results show that TaPas achieves better results in comparison to SentenceBERT. We further analyze the results of TaPas by breaking down the MRR and MAP scores based on the four reference types: cell, linked paragraph, row, and multi-row. These results are shown in Table \ref{tab:statetracking_analysis}, along with the number of samples for each reference type in the test set. We find that TaPas achieves the best overall results for row states, which also comprise the largest fraction of samples. Meanwhile, multi-row achieves a high MRR score but a low MAP score, indicating that TaPas ranks some of the correct row candidates very low. Cell and linked paragraph states are limited to a single cell within the table, but linked paragraph samples achieve noticeably better results. This is likely because the paragraph text will contain more information than a cell's text, making it easier to determine the correct reference.

\subsection{Dialogue Generation}
We adopted SacreBLEU~\cite{post-2018-call} and BERTscore~\cite{zhang2019bertscore} as the automatic evaluation metrics.
As shown in Table~\ref{tab:dialog-gen}, concatenating references can consistently improve both metrics and the collected references are necessary for generating dialogue. It can be seen that differences are more noticeable for SacreBLEU as opposed to BERTscore. This is due to the naturally similar outputs of BERTscore, where the ranking of the scores is a more reliable view of the metric. 

We conduct further error analysis and find three main types of errors as listed in Table~\ref{tab:dialog-gen-errors}: {\it incoherent}, {\it non-fluent}, and {\it unfaithful}.
As shown in Table~\ref{tab:dialog-gen-errors}, the generated response ``Alanis Nadine Morissette is a Canadian-American singer, songwriter, and actress.'' is not an appropriate response to the question. In this case, the generated response is incoherent based on the dialogue.
Sometimes the response has the correct information, but it is not a fluent sentence. One example is the generated statement ``Yes, they performed
to win the song La det swinge''.
The final primary error type is that the generated response may be unfaithful to the perceived knowledge. For example, given a paragraph mentioning several years and events in history, the generated response mentions ``1998'', while the answer should be ``2005''.

\subsection{Human Evaluation}
In addition, we conduct a human evaluation. We randomly sample 200 test samples containing previous conversation histories, human-written answers, and machine-generated answers from DialoGPT. 
For each sample, we have two Turkers provide ratings. 
We ask the Turker to evaluate the machine-generated response on three criteria:
coherence, fluency, and informativeness from a scale of 1 to 5.
Coherence measures how well the response is connected to the question and prior conversation history.
Fluency measures the use of proper English.
Informativeness measures how accurate the machine-generated response is against the human-provided ground truth response.
We provide the average ratings for each model in Table \ref{tab:human-eval}.
The model that utilizes the state tracking references achieves a better "informativeness" rating
as it is able to utilize the extra information to provide a more correct response.
It is notable however that the model with no references achieves better coherence and fluency scores.
Thus, the human evaluation demonstrates the importance and challenge for models to provide both
an accurate and articulate response.

\begin{table}[t]
    \centering
    \begin{tabular}{l|c|c|c}\toprule[1pt]
        \bf Method & \bf C  & \bf F & \bf I \\\midrule[0.5pt]
        DialoGPT-noR & \textbf{3.88} & \textbf{3.98} & 3.13 \\
        DialoGPT & 3.59 & 3.68 & \textbf{3.49} \\ \bottomrule[1pt]
    \end{tabular}
    \caption{The results of human evaluation on dialogue generation model outputs. C = Coherence, F = Fluency, I = Informativeness.}
    \label{tab:human-eval}
\end{table}

\section{Conclusion}
In this paper, we presented a novel dataset, \textsc{HybriDialogue}, for information-seeking dialogue where knowledge is grounded in both tables and text.
While previous work has combined table and text modality in the question-answering space, this has not been utilized in the dialogue setting. Our results in the various tasks demonstrate that there is still significant room for improvement and illustrate the need to build models that can adapt well to this hybrid format. In addition to the baseline tasks, future research can utilize \textsc{HybriDialogue} to explore automatic multihop question decomposition. 


\section*{Ethical Considerations}

While the dialogues in our dataset are grounded on both structured and unstructured data, they are limited to tables and text and do not cover other forms such as knowledge graphs. Additionally, the conversations are limited to discussions on single Wikipedia pages. We believe future research can expand on this for the creation of more open-ended information-seeking dialogues.

Wikipedia has extensive measures of risks and employs staff and volunteer editors to make sure Wikipedia articles meet the requirement and quality of the Wikimedia Foundation. Our data is based on Wikipedia pages, and we contain our dialogues to Wikipedia knowledge. We carefully validate the dataset collection process, and the quality of our data is carefully controlled.

The \textsc{HybriDialogue} dataset was built from the OTT-QA dataset, which is under MIT license. The authors of the OTT-QA dataset paper have allowed us to utilize the dataset within our use case.

For the dataset collection task, we required Turkers to have a HIT Approval Rate of greater than 96\% and be located in AU, CA, IE, NZ, GB, or the US. We also required workers to have had 500 HITs approved previously. Workers were shown an interface containing text input fields and navigation tools. Turkers were also given an instruction page containing a video demo and a completed example. The time to complete the task is around 5 minutes, and Turkers were paid \$1.1 per conversation, which translates to an hourly wage of \$13.2 per hour. For the human evaluation task, Turkers were paid \$0.1 per task with an estimated time of fewer than 30 seconds per task. The dataset collection protocol was approved by the IRB. We follow the user agreement on Mechanical Turk for our dataset creation, which gives us explicit consent to receive users' service in the form of data annotation in return for monetary compensation. Given our settings, the Turkers understand that their data will be utilized in machine learning research.

 We will be providing open access to our dataset for use in future research. This includes the samples of dialogues written by Mechanical Turk workers, the references that each dialogue turn is associated with, and the Wikipedia pages in which the references are located.  The dataset will be open-sourced under the MIT License.
 
 \section{Acknowledgements}
We thank the reviewers for their comments in revising this paper. This work was supported by a Google Research Award and the National Science Foundation award \#2048122. The views expressed are those of the author and do not reflect the official policy or position of the funding agencies.

\bibliography{anthology,custom}

\begin{thebibliography}{40}
\expandafter\ifx\csname natexlab\endcsname\relax\def\natexlab#1{#1}\fi

\bibitem[{Bollacker et~al.(2008)Bollacker, Evans, Paritosh, Sturge, and
  Taylor}]{bollacker2008freebase}
Kurt Bollacker, Colin Evans, Praveen Paritosh, Tim Sturge, and Jamie Taylor.
  2008.
\newblock Freebase: a collaboratively created graph database for structuring
  human knowledge.
\newblock In \emph{Proceedings of the 2008 ACM SIGMOD international conference
  on Management of data}.

\bibitem[{Brown(2020)}]{rank_bm25}
Dorian Brown. 2020.
\newblock \href {https://doi.org/10.5281/zenodo.4520057} {{Rank-BM25: A
  Collection of BM25 Algorithms in Python}}.

\bibitem[{Budzianowski et~al.(2018)Budzianowski, Wen, Tseng, Casanueva, Ultes,
  Ramadan, and Ga{\v{s}}i{\'c}}]{budzianowski-etal-2018-multiwoz}
Pawe{\l} Budzianowski, Tsung-Hsien Wen, Bo-Hsiang Tseng, I{\~n}igo Casanueva,
  Stefan Ultes, Osman Ramadan, and Milica Ga{\v{s}}i{\'c}. 2018.
\newblock \href {https://doi.org/10.18653/v1/D18-1547} {{M}ulti{WOZ} - a
  large-scale multi-domain {W}izard-of-{O}z dataset for task-oriented dialogue
  modelling}.
\newblock In \emph{Proceedings of the 2018 Conference on Empirical Methods in
  Natural Language Processing}, pages 5016--5026, Brussels, Belgium.
  Association for Computational Linguistics.

\bibitem[{Campos et~al.(2020)Campos, Otegi, Soroa, Deriu, Cieliebak, and
  Agirre}]{campos-etal-2020-doqa}
Jon~Ander Campos, Arantxa Otegi, Aitor Soroa, Jan Deriu, Mark Cieliebak, and
  Eneko Agirre. 2020.
\newblock \href {https://doi.org/10.18653/v1/2020.acl-main.652} {{D}o{QA} -
  accessing domain-specific {FAQ}s via conversational {QA}}.
\newblock In \emph{Proceedings of the 58th Annual Meeting of the Association
  for Computational Linguistics}, pages 7302--7314, Online. Association for
  Computational Linguistics.

\bibitem[{Chen et~al.(2020{\natexlab{a}})Chen, Chang, Schlinger, Wang, and
  Cohen}]{chen2020open}
Wenhu Chen, Ming-Wei Chang, Eva Schlinger, William~Yang Wang, and William~W
  Cohen. 2020{\natexlab{a}}.
\newblock Open question answering over tables and text.
\newblock In \emph{International Conference on Learning Representations}.

\bibitem[{Chen et~al.(2019)Chen, Wang, Chen, Zhang, Wang, Li, Zhou, and
  Wang}]{chen2019tabfact}
Wenhu Chen, Hongmin Wang, Jianshu Chen, Yunkai Zhang, Hong Wang, Shiyang Li,
  Xiyou Zhou, and William~Yang Wang. 2019.
\newblock Tabfact: A large-scale dataset for table-based fact verification.
\newblock In \emph{International Conference on Learning Representations}.

\bibitem[{Chen et~al.(2020{\natexlab{b}})Chen, Zha, Chen, Xiong, Wang, and
  Wang}]{chen-etal-2020-hybridqa}
Wenhu Chen, Hanwen Zha, Zhiyu Chen, Wenhan Xiong, Hong Wang, and William~Yang
  Wang. 2020{\natexlab{b}}.
\newblock \href {https://doi.org/10.18653/v1/2020.findings-emnlp.91}
  {{H}ybrid{QA}: A dataset of multi-hop question answering over tabular and
  textual data}.
\newblock In \emph{Findings of the Association for Computational Linguistics:
  EMNLP 2020}, pages 1026--1036, Online. Association for Computational
  Linguistics.

\bibitem[{Crowston(2012)}]{mechanicalturk}
Kevin Crowston. 2012.
\newblock Amazon mechanical turk: A research tool for organizations and
  information systems scholars.
\newblock In \emph{Shaping the Future of ICT Research. Methods and Approaches},
  pages 210--221, Berlin, Heidelberg. Springer Berlin Heidelberg.

\bibitem[{Devlin et~al.(2019)Devlin, Chang, Lee, and
  Toutanova}]{devlin-etal-2019-bert}
Jacob Devlin, Ming-Wei Chang, Kenton Lee, and Kristina Toutanova. 2019.
\newblock \href {https://doi.org/10.18653/v1/N19-1423} {{BERT}: Pre-training of
  deep bidirectional transformers for language understanding}.
\newblock In \emph{Proceedings of the 2019 Conference of the North {A}merican
  Chapter of the Association for Computational Linguistics: Human Language
  Technologies, Volume 1 (Long and Short Papers)}, pages 4171--4186,
  Minneapolis, Minnesota. Association for Computational Linguistics.

\bibitem[{Dinan et~al.(2018)Dinan, Roller, Shuster, Fan, Auli, and
  Weston}]{dinan2018wizard}
Emily Dinan, Stephen Roller, Kurt Shuster, Angela Fan, Michael Auli, and Jason
  Weston. 2018.
\newblock Wizard of wikipedia: Knowledge-powered conversational agents.
\newblock In \emph{International Conference on Learning Representations}.

\bibitem[{Eric et~al.(2019)Eric, Goel, Paul, Sethi, Agarwal, Gao, and
  Hakkani-Tur}]{eric2019multiwoz}
Mihail Eric, Rahul Goel, Shachi Paul, Abhishek Sethi, Sanchit Agarwal, Shuyag
  Gao, and Dilek Hakkani-Tur. 2019.
\newblock Multiwoz 2.1: Multi-domain dialogue state corrections and state
  tracking baselines.
\newblock \emph{arXiv preprint arXiv:1907.01669}.

\bibitem[{Eric et~al.(2017)Eric, Krishnan, Charette, and
  Manning}]{eric-etal-2017-key}
Mihail Eric, Lakshmi Krishnan, Francois Charette, and Christopher~D. Manning.
  2017.
\newblock \href {https://doi.org/10.18653/v1/W17-5506} {Key-value retrieval
  networks for task-oriented dialogue}.
\newblock In \emph{Proceedings of the 18th Annual {SIG}dial Meeting on
  Discourse and Dialogue}, pages 37--49, Saarbr{\"u}cken, Germany. Association
  for Computational Linguistics.

\bibitem[{Ghazvininejad et~al.(2018)Ghazvininejad, Brockett, Chang, Dolan, Gao,
  Yih, and Galley}]{ghazvininejad2018knowledge}
Marjan Ghazvininejad, Chris Brockett, Ming-Wei Chang, Bill Dolan, Jianfeng Gao,
  Wen-tau Yih, and Michel Galley. 2018.
\newblock A knowledge-grounded neural conversation model.
\newblock In \emph{Thirty-Second AAAI Conference on Artificial Intelligence}.

\bibitem[{Gopalakrishnan et~al.(2019)Gopalakrishnan, Hedayatnia, Chen,
  Gottardi, Kwatra, Venkatesh, Gabriel, and Hakkani-Tür}]{Gopalakrishnan2019}
Karthik Gopalakrishnan, Behnam Hedayatnia, Qinlang Chen, Anna Gottardi, Sanjeev
  Kwatra, Anu Venkatesh, Raefer Gabriel, and Dilek Hakkani-Tür. 2019.
\newblock \href {https://doi.org/10.21437/Interspeech.2019-3079}
  {{Topical-Chat: Towards Knowledge-Grounded Open-Domain Conversations}}.
\newblock In \emph{Proc. Interspeech 2019}, pages 1891--1895.

\bibitem[{Herzig et~al.(2020)Herzig, Nowak, M{\"u}ller, Piccinno, and
  Eisenschlos}]{herzig-etal-2020-tapas}
Jonathan Herzig, Pawel~Krzysztof Nowak, Thomas M{\"u}ller, Francesco Piccinno,
  and Julian Eisenschlos. 2020.
\newblock \href {https://doi.org/10.18653/v1/2020.acl-main.398} {{T}a{P}as:
  Weakly supervised table parsing via pre-training}.
\newblock In \emph{Proceedings of the 58th Annual Meeting of the Association
  for Computational Linguistics}, pages 4320--4333, Online. Association for
  Computational Linguistics.

\bibitem[{Iyyer et~al.(2017)Iyyer, Yih, and Chang}]{iyyer-etal-2017-search}
Mohit Iyyer, Wen-tau Yih, and Ming-Wei Chang. 2017.
\newblock \href {https://doi.org/10.18653/v1/P17-1167} {Search-based neural
  structured learning for sequential question answering}.
\newblock In \emph{Proceedings of the 55th Annual Meeting of the Association
  for Computational Linguistics (Volume 1: Long Papers)}, pages 1821--1831,
  Vancouver, Canada. Association for Computational Linguistics.

\bibitem[{Joshi et~al.(2017)Joshi, Choi, Weld, and
  Zettlemoyer}]{joshi-etal-2017-triviaqa}
Mandar Joshi, Eunsol Choi, Daniel Weld, and Luke Zettlemoyer. 2017.
\newblock \href {https://doi.org/10.18653/v1/P17-1147} {{T}rivia{QA}: A large
  scale distantly supervised challenge dataset for reading comprehension}.
\newblock In \emph{Proceedings of the 55th Annual Meeting of the Association
  for Computational Linguistics (Volume 1: Long Papers)}, pages 1601--1611,
  Vancouver, Canada. Association for Computational Linguistics.

\bibitem[{Kwiatkowski et~al.(2019)Kwiatkowski, Palomaki, Redfield, Collins,
  Parikh, Alberti, Epstein, Polosukhin, Devlin, Lee, Toutanova, Jones, Kelcey,
  Chang, Dai, Uszkoreit, Le, and Petrov}]{kwiatkowski-etal-2019-natural}
Tom Kwiatkowski, Jennimaria Palomaki, Olivia Redfield, Michael Collins, Ankur
  Parikh, Chris Alberti, Danielle Epstein, Illia Polosukhin, Jacob Devlin,
  Kenton Lee, Kristina Toutanova, Llion Jones, Matthew Kelcey, Ming-Wei Chang,
  Andrew~M. Dai, Jakob Uszkoreit, Quoc Le, and Slav Petrov. 2019.
\newblock \href {https://doi.org/10.1162/tacl_a_00276} {Natural questions: A
  benchmark for question answering research}.
\newblock \emph{Transactions of the Association for Computational Linguistics},
  7:452--466.

\bibitem[{Moon et~al.(2019)Moon, Shah, Kumar, and Subba}]{moon2019opendialkg}
Seungwhan Moon, Pararth Shah, Anuj Kumar, and Rajen Subba. 2019.
\newblock Opendialkg: Explainable conversational reasoning with attention-based
  walks over knowledge graphs.
\newblock In \emph{Proceedings of the 57th Annual Meeting of the Association
  for Computational Linguistics}.

\bibitem[{Mrk{\v{s}}i{\'c} et~al.(2017)Mrk{\v{s}}i{\'c}, {\'O}~S{\'e}aghdha,
  Wen, Thomson, and Young}]{mrksic-etal-2017-neural}
Nikola Mrk{\v{s}}i{\'c}, Diarmuid {\'O}~S{\'e}aghdha, Tsung-Hsien Wen, Blaise
  Thomson, and Steve Young. 2017.
\newblock \href {https://doi.org/10.18653/v1/P17-1163} {Neural belief tracker:
  Data-driven dialogue state tracking}.
\newblock In \emph{Proceedings of the 55th Annual Meeting of the Association
  for Computational Linguistics (Volume 1: Long Papers)}, pages 1777--1788,
  Vancouver, Canada. Association for Computational Linguistics.

\bibitem[{Pedregosa et~al.(2011)Pedregosa, Varoquaux, Gramfort, Michel,
  Thirion, Grisel, Blondel, Prettenhofer, Weiss, Dubourg, Vanderplas, Passos,
  Cournapeau, Brucher, Perrot, and Duchesnay}]{scikit-learn}
F.~Pedregosa, G.~Varoquaux, A.~Gramfort, V.~Michel, B.~Thirion, O.~Grisel,
  M.~Blondel, P.~Prettenhofer, R.~Weiss, V.~Dubourg, J.~Vanderplas, A.~Passos,
  D.~Cournapeau, M.~Brucher, M.~Perrot, and E.~Duchesnay. 2011.
\newblock Scikit-learn: Machine learning in {P}ython.
\newblock \emph{Journal of Machine Learning Research}, 12:2825--2830.

\bibitem[{Post(2018)}]{post-2018-call}
Matt Post. 2018.
\newblock \href {https://doi.org/10.18653/v1/W18-6319} {A call for clarity in
  reporting {BLEU} scores}.
\newblock In \emph{Proceedings of the Third Conference on Machine Translation:
  Research Papers}, pages 186--191, Brussels, Belgium. Association for
  Computational Linguistics.

\bibitem[{Ramadan et~al.(2018)Ramadan, Budzianowski, and
  Gasic}]{ramadan2018large}
Osman Ramadan, Pawe{\l} Budzianowski, and Milica Gasic. 2018.
\newblock Large-scale multi-domain belief tracking with knowledge sharing.
\newblock In \emph{Proceedings of the 56th Annual Meeting of the Association
  for Computational Linguistics}, volume~2, pages 432--437.

\bibitem[{Reddy et~al.(2019)Reddy, Chen, and Manning}]{reddy-etal-2019-coqa}
Siva Reddy, Danqi Chen, and Christopher~D. Manning. 2019.
\newblock \href {https://doi.org/10.1162/tacl_a_00266} {{C}o{QA}: A
  conversational question answering challenge}.
\newblock \emph{Transactions of the Association for Computational Linguistics},
  7:249--266.

\bibitem[{Reimers and
  Gurevych(2019{\natexlab{a}})}]{reimers-2019-sentence-bert}
Nils Reimers and Iryna Gurevych. 2019{\natexlab{a}}.
\newblock \href {https://arxiv.org/abs/1908.10084} {Sentence-bert: Sentence
  embeddings using siamese bert-networks}.
\newblock In \emph{Proceedings of the 2019 Conference on Empirical Methods in
  Natural Language Processing}. Association for Computational Linguistics.

\bibitem[{Reimers and
  Gurevych(2019{\natexlab{b}})}]{reimers-gurevych-2019-sentence}
Nils Reimers and Iryna Gurevych. 2019{\natexlab{b}}.
\newblock \href {https://doi.org/10.18653/v1/D19-1410} {Sentence-{BERT}:
  Sentence embeddings using {S}iamese {BERT}-networks}.
\newblock In \emph{Proceedings of the 2019 Conference on Empirical Methods in
  Natural Language Processing and the 9th International Joint Conference on
  Natural Language Processing (EMNLP-IJCNLP)}, pages 3982--3992, Hong Kong,
  China. Association for Computational Linguistics.

\bibitem[{Ren et~al.(2018)Ren, Xie, Chen, and Yu}]{ren-etal-2018-towards}
Liliang Ren, Kaige Xie, Lu~Chen, and Kai Yu. 2018.
\newblock \href {https://doi.org/10.18653/v1/D18-1299} {Towards universal
  dialogue state tracking}.
\newblock In \emph{Proceedings of the 2018 Conference on Empirical Methods in
  Natural Language Processing}, pages 2780--2786, Brussels, Belgium.
  Association for Computational Linguistics.

\bibitem[{Saeidi et~al.(2018)Saeidi, Bartolo, Lewis, Singh, Rockt{\"a}schel,
  Sheldon, Bouchard, and Riedel}]{saeidi-etal-2018-interpretation}
Marzieh Saeidi, Max Bartolo, Patrick Lewis, Sameer Singh, Tim Rockt{\"a}schel,
  Mike Sheldon, Guillaume Bouchard, and Sebastian Riedel. 2018.
\newblock \href {https://doi.org/10.18653/v1/D18-1233} {Interpretation of
  natural language rules in conversational machine reading}.
\newblock In \emph{Proceedings of the 2018 Conference on Empirical Methods in
  Natural Language Processing}, pages 2087--2097, Brussels, Belgium.
  Association for Computational Linguistics.

\bibitem[{Speer et~al.(2017)Speer, Chin, and Havasi}]{speer2017conceptnet}
Robyn Speer, Joshua Chin, and Catherine Havasi. 2017.
\newblock Conceptnet 5.5: An open multilingual graph of general knowledge.
\newblock In \emph{Thirty-first AAAI conference on artificial intelligence}.

\bibitem[{Tuan et~al.(2019)Tuan, Chen, and Lee}]{tuan2019dykgchat}
Yi-Lin Tuan, Yun-Nung Chen, and Hung-Yi Lee. 2019.
\newblock Dykgchat: Benchmarking dialogue generation grounding on dynamic
  knowledge graphs.
\newblock In \emph{Proceedings of the 2019 Conference on Empirical Methods in
  Natural Language Processing and the 9th International Joint Conference on
  Natural Language Processing (EMNLP-IJCNLP)}.

\bibitem[{Wolf et~al.(2020)Wolf, Debut, Sanh, Chaumond, Delangue, Moi, Cistac,
  Rault, Louf, Funtowicz, Davison, Shleifer, von Platen, Ma, Jernite, Plu, Xu,
  Le~Scao, Gugger, Drame, Lhoest, and Rush}]{wolf-etal-2020-transformers}
Thomas Wolf, Lysandre Debut, Victor Sanh, Julien Chaumond, Clement Delangue,
  Anthony Moi, Pierric Cistac, Tim Rault, Remi Louf, Morgan Funtowicz, Joe
  Davison, Sam Shleifer, Patrick von Platen, Clara Ma, Yacine Jernite, Julien
  Plu, Canwen Xu, Teven Le~Scao, Sylvain Gugger, Mariama Drame, Quentin Lhoest,
  and Alexander Rush. 2020.
\newblock \href {https://doi.org/10.18653/v1/2020.emnlp-demos.6} {Transformers:
  State-of-the-art natural language processing}.
\newblock In \emph{Proceedings of the 2020 Conference on Empirical Methods in
  Natural Language Processing: System Demonstrations}, pages 38--45, Online.
  Association for Computational Linguistics.

\bibitem[{Wu et~al.(2019)Wu, Guo, Zhou, Wu, Zhang, Lian, and
  Wang}]{wu2019proactive}
Wenquan Wu, Zhen Guo, Xiangyang Zhou, Hua Wu, Xiyuan Zhang, Rongzhong Lian, and
  Haifeng Wang. 2019.
\newblock Proactive human-machine conversation with explicit conversation goal.
\newblock In \emph{Proceedings of the 57th Annual Meeting of the Association
  for Computational Linguistics}, pages 3794--3804.

\bibitem[{Yagcioglu et~al.(2018)Yagcioglu, Erdem, Erdem, and
  Ikizler-Cinbis}]{yagcioglu-etal-2018-recipeqa}
Semih Yagcioglu, Aykut Erdem, Erkut Erdem, and Nazli Ikizler-Cinbis. 2018.
\newblock \href {https://doi.org/10.18653/v1/D18-1166} {{R}ecipe{QA}: A
  challenge dataset for multimodal comprehension of cooking recipes}.
\newblock In \emph{Proceedings of the 2018 Conference on Empirical Methods in
  Natural Language Processing}, pages 1358--1368, Brussels, Belgium.
  Association for Computational Linguistics.

\bibitem[{Yang et~al.(2018)Yang, Qi, Zhang, Bengio, Cohen, Salakhutdinov, and
  Manning}]{yang-etal-2018-hotpotqa}
Zhilin Yang, Peng Qi, Saizheng Zhang, Yoshua Bengio, William Cohen, Ruslan
  Salakhutdinov, and Christopher~D. Manning. 2018.
\newblock \href {https://doi.org/10.18653/v1/D18-1259} {{H}otpot{QA}: A dataset
  for diverse, explainable multi-hop question answering}.
\newblock In \emph{Proceedings of the 2018 Conference on Empirical Methods in
  Natural Language Processing}, pages 2369--2380, Brussels, Belgium.
  Association for Computational Linguistics.

\bibitem[{Zang et~al.(2020)Zang, Rastogi, Sunkara, Gupta, Zhang, and
  Chen}]{zang2020multiwoz}
Xiaoxue Zang, Abhinav Rastogi, Srinivas Sunkara, Raghav Gupta, Jianguo Zhang,
  and Jindong Chen. 2020.
\newblock Multiwoz 2.2: A dialogue dataset with additional annotation
  corrections and state tracking baselines.
\newblock In \emph{Proceedings of the 2nd Workshop on Natural Language
  Processing for Conversational AI, ACL 2020}, pages 109--117.

\bibitem[{Zhang et~al.(2019)Zhang, Kishore, Wu, Weinberger, and
  Artzi}]{zhang2019bertscore}
Tianyi Zhang, Varsha Kishore, Felix Wu, Kilian~Q Weinberger, and Yoav Artzi.
  2019.
\newblock Bertscore: Evaluating text generation with bert.
\newblock In \emph{International Conference on Learning Representations}.

\bibitem[{Zhang et~al.(2020)Zhang, Sun, Galley, Chen, Brockett, Gao, Gao, Liu,
  and Dolan}]{zhang2020dialogpt}
Yizhe Zhang, Siqi Sun, Michel Galley, Yen-Chun Chen, Chris Brockett, Xiang Gao,
  Jianfeng Gao, Jingjing Liu, and William~B Dolan. 2020.
\newblock Dialogpt: Large-scale generative pre-training for conversational
  response generation.
\newblock In \emph{Proceedings of the 58th Annual Meeting of the Association
  for Computational Linguistics: System Demonstrations}, pages 270--278.

\bibitem[{Zhou et~al.(2018{\natexlab{a}})Zhou, Young, Huang, Zhao, Xu, and
  Zhu}]{zhou2018commonsense}
Hao Zhou, Tom Young, Minlie Huang, Haizhou Zhao, Jingfang Xu, and Xiaoyan Zhu.
  2018{\natexlab{a}}.
\newblock Commonsense knowledge aware conversation generation with graph
  attention.
\newblock In \emph{IJCAI}.

\bibitem[{Zhou et~al.(2020)Zhou, Zheng, Huang, Huang, and Zhu}]{zhou2020kdconv}
Hao Zhou, Chujie Zheng, Kaili Huang, Minlie Huang, and Xiaoyan Zhu. 2020.
\newblock Kdconv: A chinese multi-domain dialogue dataset towards multi-turn
  knowledge-driven conversation.
\newblock In \emph{Proceedings of the 58th Annual Meeting of the Association
  for Computational Linguistics}, pages 7098--7108.

\bibitem[{Zhou et~al.(2018{\natexlab{b}})Zhou, Prabhumoye, and
  Black}]{zhou-etal-2018-dataset}
Kangyan Zhou, Shrimai Prabhumoye, and Alan~W Black. 2018{\natexlab{b}}.
\newblock \href {https://doi.org/10.18653/v1/D18-1076} {A dataset for document
  grounded conversations}.
\newblock In \emph{Proceedings of the 2018 Conference on Empirical Methods in
  Natural Language Processing}, pages 708--713, Brussels, Belgium. Association
  for Computational Linguistics.

\end{thebibliography}
\bibliographystyle{acl_natbib}

\appendix
\section{Appendix} \label{sec:appendix}

\subsection{Conversation Decompositions}

We counted the number and frequency of unique decompositions in our dataset, which is the selected reference sequence in a conversation. The most frequent decompositions are shown in Table \ref{tab:decompositions}.

\begin{table}[h]
\centering
\makebox[\linewidth]{
\begin{tabular}{{l}|{l}}
\toprule
Decomposition  & Count \\
\hline
$I \rightarrow T \rightarrow R \rightarrow P$    & 1419 \\
$I \rightarrow T \rightarrow R \rightarrow C$    & 733  \\
$I \rightarrow T \rightarrow R \rightarrow R$    & 290  \\
$I \rightarrow T \rightarrow R \rightarrow C \rightarrow P$  & 218  \\
$T \rightarrow R \rightarrow R \rightarrow P \rightarrow P$   & 136  \\ 
$T \rightarrow R \rightarrow P \rightarrow P$    & 116  \\
$T \rightarrow R \rightarrow C \rightarrow P$    & 116  \\
\bottomrule
 \end{tabular}
 }
\caption{Top 7 most frequent decompositions. A decomposition is defined to be the sequence of references in a given conversation. I = Intro, T = Table. R = Row, P = Linked Paragraph, C = Cell  }\label{tab:decompositions}
\end{table}

\subsection{Experimental Details}

We utilized paraphrase-distilroberta-base-v1 model with 82 million parameters provided by the SBERT library~\cite{reimers-gurevych-2019-sentence} for the SentenceBERT system state tracking experiment.
The TaPas model is built on the BERT model~\cite{devlin-etal-2019-bert}. We utilize the TaPas-base model, which correlates to the BERT-base model that contains 110 million parameters. For system state tracking evaluation, we utilize average\_precision\_score from sklearn~\cite{scikit-learn}.
For retrieval experiments, we utilized the BM25Okapi algorithm from the Rank-BM25 library \cite{rank_bm25}.
Our experiments on dialogue generation utilize DialoGPT-small in the Huggingface transformers library~\cite{wolf-etal-2020-transformers}, which contains 124 million parameters.

\end{document}